\documentclass[11pt]{article}
\usepackage[final]{nlp4call}
\usepackage{times}
\usepackage{xurl}
\usepackage{latexsym}
\usepackage{booktabs}
\usepackage[plain]{fancyref}
\usepackage{graphicx}

\title{Evaluation of really good grammatical error correction}

\author{Robert Östling$^{1}$, Katarina Gillholm$^{2}$, Murathan Kurfal{\i}$^{3}$, Marie Mattson,\\
\textbf{Mats Wir{\'e}n$^{5}$}\\
[0.5cm] 
$^{1}$Department of Linguistics / Stockholm University / {\tt robert@ling.su.se} \\
$^{2}$Department of Linguistics / Stockholm University / {\tt gillholm.ina@gmail.com} \\
$^{3}$Psychology Department / Stockholm University / {\tt murathan.kurfali@su.se} \\
$^{4}${\tt marie.mattson18@gmail.com} \\
$^{5}$Department of Linguistics / Stockholm University / {\tt mats.wiren@ling.su.se} \\
}

\date{}

\begin{document}
\maketitle
\begin{abstract}

Although rarely stated, in practice, Grammatical Error Correction (GEC) encompasses various models with distinct objectives, ranging from grammatical error detection to improving fluency. Traditional evaluation methods fail to fully capture the full range of system capabilities and objectives. Reference-based evaluations suffer from limitations in capturing the wide variety of possible correction and the biases introduced during reference creation and is prone to favor fixing local errors over overall text improvement. The emergence of large language models (LLMs) has further highlighted the shortcomings of these evaluation strategies, emphasizing the need for a paradigm shift in evaluation methodology. In the current study, we perform a comprehensive evaluation of various GEC systems using a recently published dataset of Swedish learner texts. The evaluation is performed using established evaluation metrics as well as human judges. We find that GPT-3 in a few-shot setting by far outperforms previous grammatical error correction systems for Swedish, a language comprising only 0.11\% of its training data. We also found that current evaluation methods contain undesirable biases that a human evaluation is able to reveal. We suggest using human post-editing of GEC system outputs to analyze the amount of change required to reach native-level human performance on the task, and provide a dataset annotated with human post-edits and assessments of grammaticality, fluency and meaning preservation of GEC system outputs.

\end{abstract}

\section{Introduction}

Grammatical Error Correction (GEC) is typically used in an extended sense of correcting language at multiple levels, including spelling errors, grammatical errors, word choice and idiom usage.

In the literature on evaluating GEC systems, one is rarely explicit about the purpose of the system. Following \citet{sakaguchi-etal-2016-reassessing}, we see two somewhat different objectives:

\begin{enumerate}
\item \textbf{Error detection and correction}, where grammaticality has priority over fluency. The goal is to point out individual language errors, which could ideally be fixed one by one, resulting in an acceptable text that is as close as possible to the original.

\item \textbf{General text improvement}, where fluency is on equal footing with grammaticality. The goal is to produce a text which is as close as possible to what a highly proficient writer would have produced, assuming a perfect understanding of the intended message of the original text.
\end{enumerate}

The distinction between the two objectives is less clear for writers at high proficiency levels, where changing an occasional spelling or grammar mistake typically results in a high-quality text. For a less proficient writer, a text may contain so many overlapping problems that it is difficult to identify local changes that together result in a high-quality text. If a GEC system is allowed to work directly at the level of general text improvement, its task may become significantly simpler.

The choice of objective has practical implications for how to evaluate the result. Traditional methods for GEC evaluation are reference-based, where either the GEC system output is compared to a human-created reference \citep[e.g.][]{napoles-etal-2015-ground}, or the sets of edit operations produced by the GEC system is compared to those needed to transform the original text to the human reference \citep{bryant-etal-2017-automatic}.

One important problem with reference-based evaluations is that there is typically a large and varied set of possible ways to express the same information. It is generally infeasible to approximate the full set of possibilities, although providing multiple references is a common approach in the machine translation community to alleviate this problem \citep[e.g.][]{qin-specia-2015-truly}. Results are also highly dependent on the way the references were created. \citet{freitag-etal-2020-bleu} show that biases due to ``translationese'' effects in the creation of references negatively affect the accuracy of the resulting evaluations, where interference from the source language may affect the translation to become less idiomatic in the target language. They obtained higher agreement between the automatic reference-based evaluations and human judgments by first asking human annotators to maximally paraphrase the reference sentences, to encourage diversity among the multiple references.

The references used for GEC evaluations suffer from the same bias, and often annotators are explicitly instructed to stay as close to the original text as possible \citep[Section 6.1]{Volodina2019swell}. We are aware of no GEC evaluation data which, in the style of \citet{freitag-etal-2020-bleu}, aims for a high amount of diversity in the references. This has the effect of biasing existing automatic evaluations against systems that perform paraphrasing rather than conservatively fixing individual errors.

In the related field of text summarization, \citet{Goyal2022summarization} found that automatic evaluation metrics severely underestimate the performance of large language models, further strengthening our suspicion that such powerful models necessitate a paradigm shift in evaluation methodology. In this work, we perform a comprehensive manual analysis of the output of multiple GEC systems, and point towards analysis of human post-edits as the most promising way of evaluating really good GEC systems.

\section{Related work}

\subsection{Reference-free evaluation metrics}

\citet{yoshimura-etal-2020-reference}, building on earlier work by \citet{asano-etal-2017-reference}, propose a reference-free metric named SOME that learns a weighting of grammatically, fluency, and semantic similarity scores obtained from three separate models. Interestingly, they find that tuning these weights on a dataset of human judgements results in 98\% of the weight being put on the fluency score computed as the difference between language model cross-entropy for the system output and original text. This result aligns well with the argument of \citet{sakaguchi-etal-2016-reassessing} that fluency is what GEC system ought to aim for.

\citet{islam-magnani-2021-end} go one step further and dispose of the grammaticality and semantic similarity models, and simple use language model scores combined with a filter based on string similarity measures to reject ``corrections'' that look too different from the original.\footnote{The paper somewhat unusually refers to the string similarity measures as ``syntactic similarity''.} System-level Pearson correlations with human scores are very high for all these systems: from about $0.88$ \citep{yoshimura-etal-2020-reference} to $0.98$ \citep{asano-etal-2017-reference}. However, metrics that rely heavily on language model scores do not handle semantic changes well. As \citet{islam-magnani-2021-end} point out, the SOME metric assigns a very positive score when ``He is going school.'' is corrected into  ``He He He He He He.''

% \citet{maeda-etal-2022-impara}

\subsection{Human evaluation}
\label{sec:human-evaluation}

\citet{grundkiewicz-etal-2015-human} performed a comprehensive ranking-based human evaluation of then-current GEC systems, and compared the human rankings to a number of automatic metrics. In their study, the edit distance-based MaxMatch (M$^2$) score \citep{dahlmeier-ng-2012-better} with a bias towards precision ($\beta = 0.18$) achieved the highest correlation with the human rankings. Pure machine translation metrics (BLEU, METEOR) were found to have \emph{negative} correlations with human rankings. \citet{napoles-etal-2015-ground} also performed a ranking-based human evaluation, and obtained very similar results to \citet{grundkiewicz-etal-2015-human}. However, in this case their proposed GLEU metric achieved a higher correlation with the human rankings than did the M$^2$ score. In both studies, correlations were relatively modest, with Spearman $\rho$ and Pearson $r$ in the order of 0.7--0.75 \citep{grundkiewicz-etal-2015-human} and 0.55 \citep{napoles-etal-2015-ground} for the highest-correlated metrics. \citet{10.1162/tacl_a_00470} performed a similar human evaluation of more recent Czech GEC systems, and found the agreement between human and GEC rankings to be very high, with Pearson's $r$ in the range 0.95--0.98 for GLEU, M$^2$, I-measure and their Czech adaptation of ERRANT \citep{felice-etal-2016-automatic,bryant-etal-2017-automatic}.

Rather than ranking system outputs akin to machine translation evaluation, \citet{yoshimura-etal-2020-reference} applied a range of different GEC systems to the same set of sentences and obtained human absolute scores in the dimensions of grammaticality, fluency, and meaning preservation. This allows GEC systems to be compared using each dimension individually or in combination.

\subsection{Co-evolution of systems and metrics}

As \citet{yoshimura-etal-2020-reference} demonstrated, it is possible to obtain a metric with an extremely strong correlation (up to $\rho = 0.98$) to human ratings, by optimizing the parameters of the metric with respect to human ratings. However, their best result was obtained by relying almost exclusively on the fluency score. This indicates that their data does not contain enough examples of GEC-``corrected'' sentences that are scored high by a language model, but suffer from problems like a low degree of meaning preservation. Since their data is based on the outputs of existing GEC systems, it inherits their biases towards certain types of mistakes. Most of  these systems have been developed against metrics such as GLEU and ERRANT, who tend to reward a conservative approach where precision is prioritized over recall and more substantial rewriting is discouraged. A GEC system optimized for high scores according to GLEU, ERRANT, or similar metrics, is thus less likely to suggest major changes that (when incorrect) may significantly alter the meaning of a text. The main dimension along which the performance of such systems vary is to what extent they can find and correct simple spelling, grammar or word choice errors -- the presence of which correlates strongly with a poor language model score.

We hypothesize that the co-evolution of GEC systems and their evaluation metrics has resulted in reinforcing the bias towards certain types of properties, namely a conservative approach which avoids paraphrasing. Traditionally this has not been much of an issue, since we did not have particularly good models for paraphrasing non-standard text. With the advent of large language models that excel at this task, we argue that it is time to break this circle of GEC system development and metric development.

\subsection{Swedish GEC}

We now briefly review published GEC systems for Swedish, focusing on general-coverage methods that perform automated correction. We do not cover methods specializing in specific error types, like spelling or collocations, or those that are not able to automatically suggest corrections.

Granska \citep{domeij-etal-2000-granska} is a mostly rule-based system for grammatical error detection and correction, which has later been combined with a probabilistic model \citep{bigert-knutsson2002probgranska} that uses a language model to score variants of the input sentence. Another contemporary rule-based system based on constraint grammar was developed by \citet{birn-2000-detecting}. More recently, \citet{Nyberg2022} implemented Swedish versions of the following two methods. First, the model of \citet{bryant-briscoe-2018-language}, which is based on generating variants of the input sentence and using a language model to choose the highest-scoring one. Second, using a neural machine translation model trained on artificially corupted data, generated in a fashion similar to that of \citet{grundkiewicz-etal-2019-neural}.

\section{Purpose and aims}

Most previous work on GEC evaluation has been performed using relatively limited and conservative systems, and we see a need to extend this line of work to systems based on large language models (LLMs) that are able to perform more substantial corrections than previous methods. Given the problems pointed out above with both reference-based and reference-free automated evaluation metrics, we think it is important to consider manual evaluation methods in addition to automated ones. Finally, since LLM-based systems are approaching human-level performance, we also want to include text versions created by humans in the evaluation, on equal footing with the automatic GEC systems in order to ensure a fair comparison.

Our main contributions in this work are the following:
\begin{itemize}
    \item We evaluate a set of GEC systems, belonging to diverse paradigms ranging from rule-based systems to large language models. In addition to automatic GEC systems, we also include in the evaluation paraphrases from humans following two different guidelines.
    \item We investigate how different evaluation methods, automatic and manual, compare across this range of different GEC paradigms, and point out relative strengths and weaknesses of each paradigm.
    \item We use human post-edits of GEC system outputs and human paraphrases, both to evaluate the systems and to quantify the differences between the final versions of each text after correction/paraphrasing and post-editing.
    \item We make several new resources publicly available: Human evaluations and post-edits of GEC system outputs, a novel annotation tool that was used to produce the above, and a baseline GEC system for Swedish.\footnote{Our code and data is available at \url{https://github.com/robertostling/gec-evaluation}}
\end{itemize}

\section{GEC systems}

In this work we compare a total of five Swedish GEC systems:
\begin{enumerate}
    \item \textbf{Granska}: the web API version of the rule-based system Granska \citep{domeij-etal-2000-granska}. We always accept its top suggestion for changes, but multiple suggestions that change the same span are rejected.
    \item \textbf{Nyberg MT}: the MT-based method of \citet[Section 3.3]{Nyberg2022}, training a neural machine translation system to translate artificially corrupted text back into its original form.
    \item \textbf{Nyberg LM}: the LM-based method of \citet[Section 3.4]{Nyberg2022}, using a language model to iteratively score local edits suggested by a heuristic procedure, until no further changes sufficiently improve the score.
    \item \textbf{MT}: Similar to Nyberg MT, but with more data, a modified method for introducing synthetic errors and a different architecture. Details can be found in \citet{kurfali-ostling-2023-distantly}.
    \item \textbf{GPT-3}: OpenAI's (\texttt{text-davinci-002}) model \citep{brown2020gpt3} through their public API. We use a two-shot prompt, with authentic student sentences taken from the CrossCheck corpus\footnote{\url{https://www.csc.kth.se/tcs/projects/xcheck/korpus.html}} and manually corrected by us. The prompt is entirely in Swedish, and is identical for all processed sentences.
\end{enumerate}
We also consider the dummy baseline \textbf{Uncorrected}, which simply leaves the text unchanged, and the following three human-corrected versions:
\begin{enumerate}
    \item \textbf{Human minimal}: human-normalized sentences from the SweLL project \citep{Volodina2019swell}. Annotators were asked to perform minimal edits in order to produce a grammatically correct sentence while trying to preserve the meaning; for normalization guidelines, see \citet{Rudebeck-et-al-2021}. The annotators had access to the full context for each sentence.
    \item \textbf{Human fluent}: human-normalized sentences produced by having a native Swedish annotator, different from the other annotators in the project, edit the \textbf{Human minimal} sentences to achieve native-like fluency while staying as close as possible to the original.
    \item \textbf{Human free}: human-normalized sentences produced by having a native Swedish annotator, different from the other annotators in the project, edit the \textbf{Human minimal} sentences to achieve native-like fluency while being encouraged to change the sentence as much as needed to achieve the most idiomatic way of expressing the given meaning.
\end{enumerate}

\section{Data and annotation}
\label{sec:data}

We use Swedish data from the SweLL project \citep{Volodina2019swell}, which consists of 502 student texts collected from different levels of L2 Swedish education. The texts are annotated with an approximate CEFR level, and have been manually normalized by minimally editing them into a grammatically correct version. We use the sentence segmentations and the division into a test and a development set from \citet{Nyberg2022}. For the systems \textbf{Nyberg MT} and \textbf{Nyberg LM}, we report evaluation results from \citet{Nyberg2022}.

Two independent annotators, both co-authors of this paper and native Swedish speakers, were tasked with performing the following procedure for each corrected sentence in two pilot datasets, containing ten sentences each:
\begin{enumerate}
    \item In a text box, read the system's output and if necessary modify it to reach the level of a native writer, considering both fluency and grammaticality. The annotators are instructed to perform the minimum amount of editing to reach this goal.
    \item When the (possibly) edited system output is submitted, the existing human-normalized reference from the SweLL data is shown (\textbf{Human minimal}), and the annotator is asked to confirm whether the meaning of the edited sentence matches the reference. If the annotator thinks otherwise, the human reference is hidden again and the tool returns to step 1.
    \item When the edited system output is accepted, the annotator is shown the student sentence, the non-edited system output, and the SweLL reference. Then the annotator chooses a score on a 4-level Likert scale (or ``other'') for the three dimensions of grammaticality, fluency and meaning preservation. We follow \citet{yoshimura-etal-2020-reference} for the definition of the scales for each of these three dimensions. Our annotation tool and full guidelines are publictly available.\footnote{\url{https://github.com/robertostling/gec-evaluation/tree/main/annotator}}
\end{enumerate}

\begin{table}[tb]
    \begin{tabular}{lrrr}
        \toprule
                    & Gramm. & Fluency & Meaning \\
        \midrule
         Round 1    & 0.45 & 0.69 & 0.51 \\
         Round 2    & 0.84 & 0.81 & 0.71 \\
         \bottomrule
    \end{tabular}
    \caption{Quadratically weighted kappa (QWK) between the two annotators during the two-round pilot phase, for grammaticality, fluency and meaning preservation.}
    \label{tab:iaa}
\end{table}

Sentences were randomized from a pool containing the outputs of all systems under evaluation, and a custom-made annotation tool was used for the task. After discussions between the annotators and after the two pilot datasets, the level of agreement was high enough (QWK in the range 0.71--0.84) to allow one annotator to continue annotating the full data. Details are presented in \Fref{tab:iaa}. The full dataset was subsampled from the development set of \citet{Nyberg2022} and contains 64 sentences each from CEFR proficiency levels A, B and C, for a total of 192 sentences. Each sentence has been processed by three GEC systems and two human paraphrasers, for a total of $192 \times 5 = 960$ output sentences with post-edits and scores.

As the final result, we have for each sentence produced by a GEC system: (a) scores for grammaticality, fluency, and meaning preservation; (b) a post-edited version of the output with the minimal edits required to obtain maximum scores on grammaticality, fluency and meaning preservation.

In order to achieve comparability with previous work on Swedish GEC, we adapted the sentence-level train/test split of \citet{Nyberg2022}, and continued performing all analysis on the sentence level. The only instance where a wider context is used, is for the minimal normalization from the SweLL project data. These normalized versions are used as references, but since GEC systems have access to less context than the reference was based on, models that are able to take longer contexts into account are put at a disadvantage during evaluation. In several cases, it is even impossible to correctly and unambiguously interpret the sentence without further context. Only the human corrections, which are directly or indirectly based on the full context, do not suffer from this problem. We have manually inspected all cases of ``moderate'' or ``substantial'' differences in meaning for the annotations of  the best-performing GEC system's (GPT-3) output, and found that 8 out of 28 (29\%) such sentences require further context. While this is a methodological problem to be addressed in future work, we see that the impact of this problem is limited.

\section{Results}

\subsection{Automatic evaluation metrics}

\begin{table}[tb]
    \centering
    \begin{tabular}{lllll}
        \toprule
                & \multicolumn{4}{c}{CEFR level} \\
         System & All & A & B & C \\
         \midrule
         Uncorrected & 0.44 & 0.29 & 0.17 & 0.53 \\
         Granska & 0.47 & 0.35 & 0.24 & 0.55 \\
         Nyberg MT & 0.51 & 0.42 & 0.30 & 0.58 \\
         Nyberg LM & 0.52 & 0.42 & 0.32 & 0.58 \\
         MT & 0.57 & 0.48 & 0.38 & 0.63 \\
         %MT-large (ours)& --  & 0.49  & 0.39 & 0.63 \\
         GPT-3 & 0.63 & 0.60 & 0.52 & 0.65 \\
         Human minimal & 1.0 & 1.0 & 1.0 & 1.0 \\
         \bottomrule
    \end{tabular}
    \caption{Reference based evaluation: GLEU scores on the test set of \citet{Nyberg2022}.}
    \label{tab:nyberg_gleu}
\end{table}

\begin{table}[tb]
    \centering
    \begin{tabular}{lllll}
        \toprule
                & \multicolumn{4}{c}{CEFR level} \\
         System & All & A & B & C \\
         \midrule
         Uncorrected & 0 & 0 & 0 & 0 \\
         Granska & 0.03 & 0.08 & 0.11 & -0.01 \\
         MT & 0.51 & 0.57 & 0.68 & 0.43 \\
         GPT-3 & 0.69 & 0.70 & 0.83 & 0.65 \\
         Human minimal & 0.68 & 0.67 & 0.77 & 0.65 \\
         \bottomrule
    \end{tabular}
    \caption{Reference-free evaluation: normalized scribendi scores on the test set of \citet{Nyberg2022}.}
    \label{tab:nyberg_scribendi}
\end{table}

\Fref{tab:nyberg_gleu} shows the performance of each system using the reference-based GLEU metric, while \Fref{tab:nyberg_scribendi} contains the corresponding evaluation using the reference-free Scribendi score \citep{islam-magnani-2021-end}.\footnote{Since figures for \textbf{Nyberg MT} and \textbf{Nyberg LM} are taken from \citet{Nyberg2022}, which only reports GLEU, they are missing from \Fref{tab:nyberg_scribendi}.} Both metrics yield the same ranking of the systems: GPT-3 scores best, followed by the NMT systems, followed in turn by the rule-based system. However, the relative differences between the systems differ considerably between the metrics. In particular, for the Scribendi score (\Fref{tab:nyberg_scribendi}) we see a very sharp divide between the neural and the non-neural systems. For all different levels, GPT-3 in fact scores higher than the human reference, even though its output contains a substantial amount of errors (as shown in the human evaluation, see \Fref{tab:human_evaluations}). This is not very surprising, since the Scribendi score mainly represents the number of sentences where the Swedish GPT-SW3 model \citep{ekgren-etal-2022-lessons} assigns a higher score to the system output than to the original sentence, and the GPT-3 output was obtained from the same family of language models.

\subsection{Human evaluation}

\begin{table}[tb]
    \centering
    \begin{tabular}{lllll}
        \toprule
                & \multicolumn{4}{c}{CEFR level} \\
         System & All & A & B & C \\
         \midrule
         \multicolumn{5}{c}{\bf Grammiticality} \\
         \midrule
         Granska            & 3.0 & 3.1 & 2.5 & 3.3 \\
         MT                 & 3.3 & 3.4 & 3.0 & 3.5 \\
         GPT-3              & 3.7 & 3.8 & 3.6 & 3.8 \\
         Human fluent       & 3.9 & 3.9 & 3.8 & 3.9 \\
         Human free         & 3.9 & 3.9 & 3.9 & 3.8 \\
         \midrule
         \multicolumn{5}{c}{\bf Fluency} \\
         \midrule
         Granska            & 2.8 & 3.0 & 2.3 & 3.2 \\
         MT                 & 3.1 & 3.2 & 2.7 & 3.3 \\
         GPT-3              & 3.6 & 3.7 & 3.4 & 3.7 \\
         Human fluent       & 3.8 & 3.8 & 3.7 & 3.8 \\
         Human free         & 3.8 & 3.8 & 3.9 & 3.8 \\
         \midrule
         \multicolumn{5}{c}{\bf Meaning preservation} \\
         \midrule
         Granska            & 3.5 & 3.5 & 3.2 & 3.7 \\
         MT                 & 3.4 & 3.5 & 3.1 & 3.6 \\
         GPT-3              & 3.4 & 3.6 & 3.1 & 3.6 \\
         Human fluent       & 3.9 & 4.0 & 3.9 & 3.8 \\
         Human free         & 3.8 & 3.8 & 3.8 & 3.8 \\
         \midrule
         \multicolumn{5}{c}{\bf Normalized Levenshtein distance (NLD)} \\
         \midrule
         Granska            & 0.126 & 0.119 & 0.180 & 0.079 \\
         MT                 & 0.113 & 0.095 & 0.158 & 0.087 \\
         GPT-3              & 0.076 & 0.068 & 0.112 & 0.050 \\
         Human fluent       & 0.034 & 0.034 & 0.045 & 0.022 \\
         Human free         & 0.029 & 0.030 & 0.034 & 0.025 \\
         \bottomrule
    \end{tabular}
    \caption{Human evaluation: mean score per system and assessment dimension.
    %Note that normalized Levenshtein distance (NLD) is with respect to the human post-edited version of a system's output, which was produced at the same time as the grammaticality, fluency and meaning preservation ratings.
    Higher is better for all assessments (range: 1--4), while lower is better for NLD.
    }
    \label{tab:human_evaluations}
\end{table}

\begin{table*}[tb]
    \centering
    \begin{tabular}{lrrrrr}
        \toprule
        System & Identical & Minor & Moderate & Substantial & Other \\
        \midrule
         Granska    & 125 & 34 & 11 & 13 &  9\\
         MT         & 122 & 35 & 19 & 13 & 3 \\
         GPT-3      & 126 & 36 & 14 & 14 & 2 \\
         Human fluent     & 176 & 11 &  3 &  1 & 1 \\
         Human free       & 160 & 26 & 5 &  0 & 1 \\
         \bottomrule
    \end{tabular}
    \caption{Human evaluation: actual distribution of meaning preservation scores. The mean over each row in this table is summarized in the \emph{All} column under \emph{Meaning preservation} in \Fref{tab:human_evaluations}.}
    \label{tab:meaning_preservation}
\end{table*}

The result of the human evaluation is summarized in \Fref{tab:human_evaluations}. As for grammaticality and fluency, the ranking of the GEC systems is identical to that of the automatic metrics. In both cases, GPT-3 performs at or near human levels. The MT-based system follows, at a considerable distance, and the rule-based system scores last. The trend is identical across CEFR proficiency levels. For meaning preservation, the situation is different. Here, there are no major differences between systems, all of them consistently perform below human levels. The gap to the human paraphrases is highest for the B level sentences, which contain the most errors and are generally the most difficult to correct.
This is presented in more detail in \Fref{tab:meaning_preservation}, where the frequency of each individual score is given. While the human corrections nearly always are classified as having no or minor differences, all automatic systems have significant numbers of moderate and substantial differences. Even given the fact that about 30\% of these divergences are due to insufficient context (see \Fref{sec:data}), the difference is large enough to indicate that all GEC systems have problems producing corrections with adequate semantics.

\subsection{A tree of corrections}

\begin{figure*}[tb]
    \includegraphics[width=0.96\textwidth]{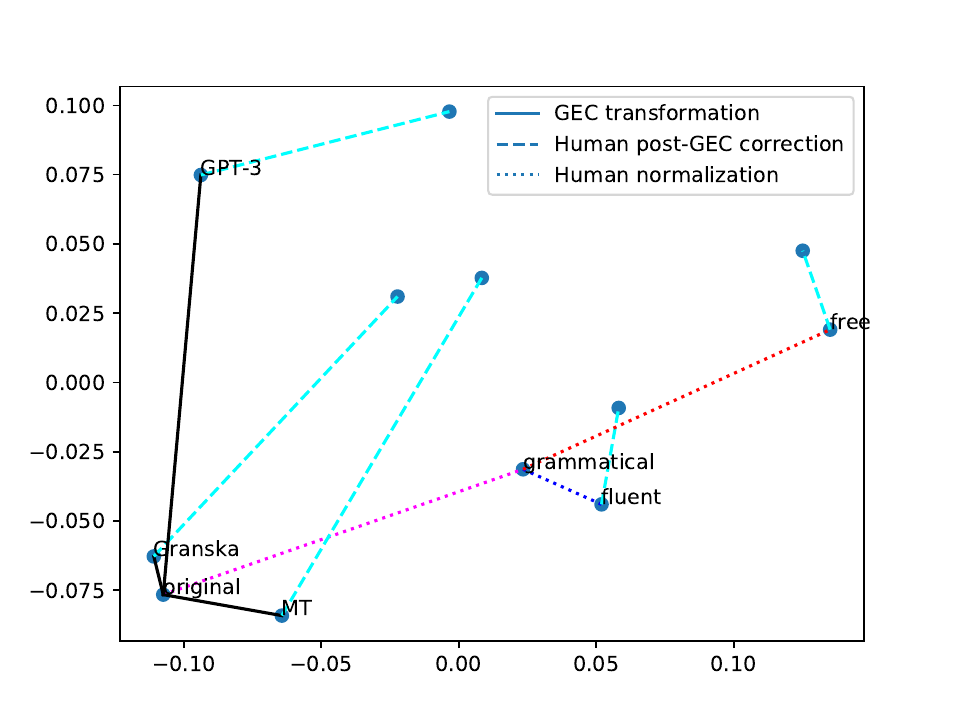}
    \caption{Multidimensional scaling view of all text versions, using normalized Levenshtein distance. The resulting graph is a tree, with its root at the ``Original'' node that represents the original text of the students. Edges represent transformations made by either computers (solid black lines) or humans (all other line types). Unlabeled leaf nodes represent human post-edits to their parent nodes in order to achieve full grammaticality, fluency and meaning preservation. Each human annotator is represented by a different collor.}
    \label{fig:mds}
\end{figure*}

In this project, we have produced a total of ten different versions of each sentence (three GEC systems and two humans, each with a post-edit), in addition to the original and its minimal correction from the SweLL project data. We visualize this by computing the normalized character-level Levenshtein distance between each pair among the twelve versions of each sentence, then using multidimensional scaling \citep{kruskal1964mds} as implemented by \citet{scikit-learn}.

In \Fref{fig:mds}, we show the result as a tree starting at the original student text (``original''), through its immediate transformations by each of the GEC systems (solid lines) and the minimal human correction (``grammatical''), followed by the human rewrites for fluency (``fluent'' and ``free'') to the post-edited versions (dashed lines). This figure complements the bottom part of \Fref{tab:human_evaluations}, which gives the distances of the post-editing edges.

It is clear that GPT-3 by far performs the most extensive changes to the text among the GEC systems. Still, the amount of post-editing required is much higher than the human-corrected versions. To some extent this can be explained by the fact that the ``fluent'' and ``free'' human corrections indirectly have access to a wider context through the SweLL ``grammatical'' correction, but as was shown in \Fref{sec:data} the size of  this effect is limited.

We can also see in \Fref{fig:mds} that although the post-edits (leaf nodes) converge somewhat, they still produce rather different versions. According to the human annotator, all of the leaf nodes represent perfect corrections and none of them should be penalized in an evaluation. From the figure, we get an impression of the space of acceptable versions of the text in relation to the space of unacceptable versions.

\section{Discussion}

We have compared a diverse set of GEC systems (rule-based, MT-based, LLM-based, human), using a diverse set of metrics (reference-based, reference-free, human scoring, human post-editing). This allows us to make some general observations on the problem of GEC evaluation.

\subsection{Metrics}

The reference-free metric (\Fref{tab:nyberg_scribendi}) shows that the neural methods (MT, GPT-3) achieve very high scores, with GPT-3 scoring above or on par with the minimal human correction. The rule-based Granska system, however, achieves scores comparable to the baseline of leaving the text uncorrected. By comparing with the human evaluation (\Fref{tab:human_evaluations}), we see considerable differences. To begin with, the human evaluation demonstrates that all GEC systems are clearly below human-level preformance. We also see that while MT is clearly better than Granska with respect to grammaticality, fluency and post-edit distance, the gap is by no means as large as suggested by the Scribendi score.

As seen in \Fref{tab:nyberg_gleu}, the reference-based GLEU suffers from ceiling effects, in particular for data from the most proficient (level C) student group. The GLEU scores are nearly identical (0.63 and 0.65) for the MT and GPT-3 systems, but in terms of grammaticality, fluency and post-edit distance there is a clear difference.

We argue that post-edit distance, when affordable, is perhaps the fairest single-dimensional GEC evaluation metric. In this work we quantify the edit distance using normalized character-level Levenshtein distance, but other edit distances that better model moved text segments may be more appropriate.

\subsection{Methodological issues}

An important lesson for future work concerns the importance of producing test sets with sufficiently long context, preferably whole documents. This would allow models with long context windows to demonstrate their full potential, and give a fair comparison to humans and to other computational models.

We also note that our choice of relying on the minimal corrections from the SweLL data as gold standard is sometimes problematic, since multiple corrections with different semantics can be plausible. In our work, we used these minimal corrections as a basis for the other (``fluent'' and ``free'') human corrections, which has the effect of reducing the diversity among the human corrections. If multiple human corrections from the original text were to be performed, we would also recommend annotating which cases are truly ambiguous even to a human. In addition, it would be helpful to include annotations of the grammaticality and fluency of the original sentence, for reference.

\subsection{Summary and future work}

In conclusion, we show that with the advent of large language models, Swedish GEC has made enormous progress compared to early work. One of these models, GPT-3, produces corrections with human-like grammaticality and fluency. However, in the critical aspect of semantic accuracy, we see little improvment compared to other types of models.

For evaluating GEC systems, we demonstrate that different types of automatic evaluation metrics display different biases with respect to different types of GEC systems. Reference-free metrics favor neural systems, even over human corrections, while reference-based metrics struggle to differentiate systems at high proficiency levels.

In this work, we used simple Normalized Levenshtein distance to quantify the differences between post-edited corrections. For future work, we believe that a more thorough analysis of these differences would provide valuable insights into the weaknesses remaining even in very strong GEC systems. This could be done manually, or in some cases automatically similar to \citet{felice-etal-2016-automatic}.

Given the strong and rapidly improving ability of LLMs to handle extended contexts, we also see a need to perform future evaluations on longer segments of texts, including entire documents. However, working at the document level requires considerable adaptations of most existing evaluation methods, which would be another interesting direction of future work.

\bibliographystyle{acl_natbib}
\bibliography{nlp4call}

\end{document}